\documentclass[conference]{IEEEtran}
\IEEEoverridecommandlockouts
\usepackage{cite}
\usepackage{amsmath,amssymb,amsfonts}
\usepackage{algorithmic}
\usepackage{graphicx}
\usepackage{textcomp}
\usepackage{xcolor}
\usepackage{listings}
\usepackage{url}
\usepackage[acronym]{glossaries}
\def\BibTeX{{\rm B\kern-.05em{\sc i\kern-.025em b}\kern-.08em
    T\kern-.1667em\lower.7ex\hbox{E}\kern-.125emX}}

\newacronym{AML}{AML}{AutomationML}
\newacronym{AR}{AR}{Application Recommendation}
\newacronym{LLM}{LLM}{Large Language Model}
\newacronym{OWL}{OWL}{Web Ontology Language}
\newacronym{RDF}{RDF}{Resource Description Framework}
\newacronym{RML}{RML}{RDF Mapping Language}
\newacronym{SHACL}{SHACL}{Shapes Constraint Language}

\begin{document}

\title{Automated Validation of Textual Constraints Against AutomationML via LLMs and SHACL
\thanks{This research article is funded by dtec.bw – Digitalization and Technology Research Center of the Bundeswehr as part of the project \textit{ProMoDi}. dtec.bw is funded by the European Union – NextGenerationEU.}
} 
\author{
\IEEEauthorblockN{
    Tom Westermann\IEEEauthorrefmark{1},
    Aljosha Köcher\IEEEauthorrefmark{1},
    Felix Gehlhoff\IEEEauthorrefmark{1},
}
\IEEEauthorblockA{
\IEEEauthorrefmark{1}Institute of Automation\\
Helmut Schmidt University, Hamburg, Germany\\
Email: \{tom.westermann, aljosha.koecher, felix.gehlhoff\}@hsu-hh.de\\}
}

\maketitle

\begin{abstract}
AutomationML (AML) enables standardized data exchange in engineering, yet existing recommendations for proper AML modeling are typically formulated as informal and textual constraints. 
These constraints cannot be validated automatically within AML itself.
This work-in-progress paper introduces a pipeline to formalize and verify such constraints. First, AML models are mapped to OWL ontologies via RML and SPARQL. 
In addition, a Large Language Model (LLM) translates textual rules into SHACL constraints, which are then validated against the previously generated AML ontology. 
Finally, SHACL validation results are automatically interpreted in natural language.
The approach is demonstrated on a sample AML recommendation. Results show that even complex modeling rules can be semi-automatically checked --- without requiring users to understand formal methods or ontology technologies.
\end{abstract}


\section{Introduction}\label{sec:introduction}
Engineering projects often involve the exchange of data across heterogeneous systems and tools, making standardized data formats essential. As an XML-based data exchange format, \gls{AML} addresses this need\cite{IEC62714-1:2018}. It structures engineering data through semantic constructs, enabling consistent representation of components and interfaces.

In order to support diverse application domains and company-specific requirements, \gls{AML} allows for extensions via reusable libraries. While \gls{AML} supports extensibility through libraries --- often specified in \glspl{AR} --- the correct use of these library elements is typically described in informal documentation written in natural language.
Additionally, \gls{AML} does not offer a suitable validation mechanism\cite{Westermann2025}. As a result, formal validation of \gls{AR}-specific modeling rules within \gls{AML} models remains challenging.

Recent work has demonstrated that transforming \gls{AML} to \gls{OWL} and validating it using \gls{SHACL} is a feasible approach for enforcing such constraints\cite{Westermann2025}. 
\gls{OWL} allows \gls{AML} data to be represented in a semantic, machine-readable format compatible with the \gls{RDF} data model. The benefit of such a conversion from \gls{AML} to \gls{OWL}is that \gls{SHACL} can then be used to define and execute validation rules over RDF data — enabling automated checking of structural and semantic correctness.
However, creating \gls{SHACL} shapes is a time consuming process that requires thorough domain knowledge and ontological expertise.

We therefore propose leveraging \glspl{LLM} to generate \gls{SHACL} shapes directly from textual constraints. The resulting \gls{SHACL} constraints enable automated validation of \gls{AML} models, bridging the gap between informal documentation and formal model verification.

The paper is structured as follows: Section~\ref{sec:relatedwork} reviews related work, Section~\ref{sec:approach} describes the proposed approach, and 
Section~\ref{sec:validation} presents an example validation use case where automatically generated \gls{SHACL} shapes are applied to a sample \gls{AML} model. 
Section \ref{sec:conclusion} summarizes the paper and outlines future work.
\section{Related Work}\label{sec:relatedwork}
\gls{AML} is an XML-based data format used to represent engineering information. 
However, \gls{SHACL} operates exclusively on \gls{RDF} data. 
This necessitates a translation from \gls{AML} to \gls{OWL}/\gls{RDF} before \gls{SHACL}-based validation can be applied.
Several approaches have been proposed to bridge this gap. The earliest effort to map \gls{AML} to \gls{RDF} was introduced in \cite{Runde2009}. 
Subsequent works \cite{GrangelGonzalez.2016,GrangelGonzalez.2018,OlgaKovalenko.2018} expanded the scope of \gls{AML} ontologies, focusing primarily on data integration tasks. However, these contributions have not addressed the use of \gls{SHACL} for validation purposes.
A recent development is the publication of a new \gls{AML} ontology alongside a declarative \gls{RML} mapping \cite{Westermann2025}. This enables automated and comprehensive translation of \gls{AML} to \gls{RDF}, making it a promising candidate for \gls{SHACL}-based validation workflows. 
It further showed that \gls{SHACL} allows new validation use cases that are not possible with XML-based tools, such as XSD or Schematron.

While these transformation approaches facilitate \gls{RDF}-based handling of \gls{AML}, none directly addresses validation to ensure correct \gls{AML} usage. 
An earlier study explored automated plant validation using CAEX data, a core component of \gls{AML} \cite{L.Abele.2015}. 
However, it relied heavily on manual modeling of validation rules and predates the availability of both \gls{SHACL} and \glspl{LLM}, thus limiting its potential for automated use.

Besides these approaches to transform \gls{AML}, there is related work focusing on the automated generation of \gls{SHACL} shapes in order to validate existing \gls{RDF} data in general.

Laurenzi et al.\cite{LMM_AnLLMAidedEnterpriseKnowledge_2024} present a six-step process in which \glspl{LLM} are applied to generate ontology schemas, \gls{RDF} data, SPARQL queries, and \gls{SHACL} shapes based on textual input. Their results demonstrate that \glspl{LLM} can produce valid \gls{SHACL} shapes from both ontology definitions and example instances, thereby reducing manual effort in constraint engineering. 
Although not tailored to \gls{AML}, their findings highlight the potential of \glspl{LLM} to automate \gls{SHACL} shape creation across domains.

Another recent contribution is the \emph{R2[RML]-ChatGPT Framework} in \cite{RaO_R2RMLChatGPTFramework_2024}, which integrates \glspl{LLM} into a mapping process for linked data generation. 
The framework leverages ChatGPT to automatically generate \gls{SHACL} shapes, SPARQL queries, and \gls{RDF} sample data based on concepts extracted from declarative mappings. 
A structured prompting and post-processing pipeline is used to ensure syntactic correctness, while semantic accuracy is evaluated against established ontologies such as \emph{FOAF} and \emph{PROV-O}. 
Their experiments show that \glspl{LLM} can produce syntactically correct \gls{SHACL} shapes with minimal intervention. Similar to our approach, \gls{SHACL} shapes are generated from a declarative mapping. However, the shapes in \cite{RaO_R2RMLChatGPTFramework_2024} are rather simple compared to the complex structures that can be required for \gls{AML} constraints.
\section{Approach}
\label{sec:approach}

\begin{figure*}[htb]
    \centering
    \includegraphics[width=0.8\textwidth]{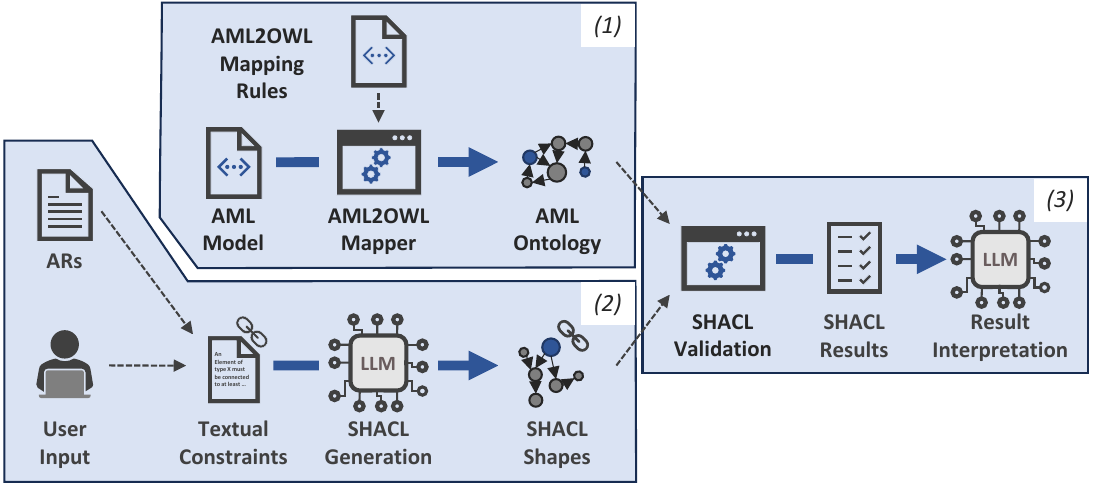}
    \caption{Overview of the AML validation approach. After generating an ontology for an \gls{AML} model \emph{(1)}, textual constraints are converted into SHACL shapes \emph{(2)} before these shapes are validated against the ontology and the results are automatically interpreted \emph{(3)}.}
    \label{fig:approachOverview}
\end{figure*}

To enable formal verification of industrial automation models in \gls{AML}, we propose a pipeline consisting of three steps. An overview of this pipeline can be seen in Figure~\ref{fig:approachOverview}.
First, \gls{AML} models are translated into an ontological representation using the declarative mapping approach presented in \cite{Westermann2025} (see Figure~\ref{fig:approachOverview}a). 
The second step of the pipeline (see Figure~\ref{fig:approachOverview}b) is the generation of \gls{SHACL} shapes from textual inputs or documents such as \glspl{AR}.
And lastly, Step 3 (see Figure~\ref{fig:approachOverview}c) consists of the actual validation of the generated ontology with the generated \gls{SHACL} shapes. All three steps are presented in more detail in the following subsections.

\subsection{Mapping an AML Model to an OWL Ontology}

The first step, described in detail in our prior work \cite{Westermann2025}, is to transform an \gls{AML} model into an OWL representation that can be used for automated validation in subsequent steps. 
To automatically transform from \gls{AML} to OWL, a so-called \emph{AML Ontology}, i.e., an OWL ontology of the \gls{AML} meta model was previously defined\footnote{\url{https://github.com/hsu-aut/IndustrialStandard-ODP-AutomationML}}.
In addition, declarative mapping rules were defined in \gls{RML} to transform arbitrary \gls{AML} models -- including all types of \gls{AML} classes as well as instances -- into this ontology. 
The mapping yields an \gls{OWL} ontology that closely mirrors the original \gls{AML} structure. For a more in-depth comparison of \gls{AML} and \gls{OWL} and a detailed description of the mapping, please refer to \cite{Westermann2025}.

\subsection{SHACL Shape Generation}
To validate constraints against the \gls{AML} ontology created in Step 1, \gls{SHACL} is used.
Constraints against \gls{AML} models are typically not available as \gls{SHACL} shapes, but rather defined in natural language, making it hard to validate constraints automatically.
One source of such constraints are \glspl{AR}, which provide official expert guidelines on how to create \emph{good} \gls{AML} models. For example, an \gls{AR} could require all InterfaceClass definitions for communication interfaces to follow a naming convention and to include predefined AttributeTypes for compatibility checking.
Furthermore, users may define custom constraints, e.g., when defining requirements against an \gls{AML} model to be created.

To generate SHACL shapes for textual constraints, an \gls{LLM} is employed. A few-shot prompting technique is used and the prompt consists of the following main parts:
\begin{itemize}
    \item \emph{Ontology context}: A condensed description of the \gls{AML} ontology, focusing on how certain \gls{AML} elements are represented in \gls{OWL}.
    \item \emph{Relevant libraries}: Libraries (typically defined in \glspl{AR}) about the \gls{AML} model under consideration in XML syntax.
    \item \emph{Examples}: Example textual constraints with correct \gls{SHACL} shapes to serve as guidance for the \gls{LLM}.
    \item \emph{Constraints}: Plain text descriptions of the constraints to generate \gls{SHACL} shapes for.
\end{itemize}
In addition to these main parts, additional commands to correctly build identifiers, define prefixes, and return output without any explanations or markdown are included in the prompt.
The result of this step is a \gls{SHACL} shape for each of the constraints passed in the prompt. These individual shapes are combined into one \gls{RDF} shape model.

\subsection{Constraint Validation and Interpretation}
The last step (see Figure~\ref{fig:approachOverview}c) contains the actual validation.
The generated \gls{SHACL} shape model is validated against the generated ontology of Step 1. 
This process produces a standardized \gls{SHACL} report highlighting possible constraint violations. 
As these violations are also returned as \gls{RDF}, they are rather difficult to interpret for users without \gls{RDF} and \gls{SHACL} expertise.
To assist in understanding violations, an \gls{LLM} is used to interpret the \gls{SHACL} results and provide additional explanations in natural language. 
By supplying the generated ontology, the \gls{SHACL} report as well as the \glspl{AR} libraries as context, the \gls{LLM} can generate human-readable explanations, which help to find modeling errors in the \gls{AML} model.

\smallskip

The approach is implemented as a prototypical implementation that can be used as a command-line tool\footnote{https://github.com/hsu-aut/aml-shacl}. The tool accepts \glspl{AR}, an \gls{AML} file to be checked and a text file with constraints as inputs. It returns validation results as well as a natural-language interpretation of these results.
\section{Validation}\label{sec:validation}
\subsection{Example Use Case}
To demonstrate the use of \gls{SHACL} for validating \gls{AML} models, we reference an example from the \gls{AR} \emph{Automation Project Configuration} (AR APC)\cite{ARAPC}. 
This \gls{AR} defines standardized RoleClasses and InterfaceClasses to model automation projects, along with rules that their correct usage. These elements are designed for reuse across projects and help ensure consistency in \gls{AML} modeling of automation projects.

As a case study, we focus on the connection between two core elements: \emph{Subnets} and \emph{Nodes}.
A \emph{Subnet} models a communication network, such as Ethernet or PROFIBUS, and is responsible for storing relevant properties and functionalities.
A \emph{Node} describes a networked device and contains networking information like logical addresses and subnet masks.
The \emph{LogicalEndPoint} InterfaceClass is used to connect \emph{Subnets} and \emph{Nodes}. 

\begin{figure}[htbp]
    \includegraphics[width=0.9\linewidth]{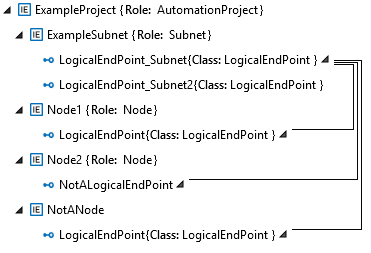}
    \caption{Excerpt of an \gls{AML} model for an automation project with a subnet that was defined in partial violation of the AR APC's rules.}
    \label{fig:automationMLExample}
\end{figure}

The AR APC defines specific rules for this connection:
\begin{itemize}
    \item \emph{Rule 1 -- LogicalEndPoint Cardinality}:
    
    Both Subnet and Node elements must include exactly one LogicalEndPoint interface.
    \item \emph{Rule 2 -- Interface-to-interface connections}:
    
    A LogicalEndPoint may only be connected to another LogicalEndPoint.
    \item \emph{Rule 3 -- Permitted connection direction}:
    
    A Subnet’s LogicalEndPoint may only be connected to the LogicalEndPoint of a Node.
\end{itemize}
Figure \ref{fig:automationMLExample} illustrates a deliberately incorrect model violating each of these rules:
\begin{itemize}
    \item The ExampleSubnet includes two LogicalEndPoint interfaces (violates Rule 1).
    \item The LogicalEndPoint connects to an interface that is not a LogicalEndPoint (violates Rule 2).
    \item The LogicalEndPoint connects to a LogicalEndPoint that is not part of a Node (violates Rule 3).
\end{itemize}
These domain-specific rules can be formalized using \gls{SHACL} to enable automated model validation. 
In the following, we first present manually created \gls{SHACL} shapes that encode these constraints, and then compare them to shapes generated with our automated pipeline. 

To investigate whether these shapes can be derived automatically, each of the three rules above was supplied to an \gls{LLM}. For this validation, we used GPT in version gpt-4.1-2025-04-14.
This included a brief synopsis of the AutomationML ontology and relevant excerpts from the AR APC rule libraries. These materials provided the LLM with the structural background, normative constraints, and concrete shape patterns necessary to generate SHACL representations of the rules.

The \gls{LLM} produced candidate shapes that were largely correct but still required light post-editing by \gls{AML} experts. The manual effort was confined to fine-tuning target declarations or adding small filters; the core constraint logic was almost always in place. A brief rule-by-rule assessment is given below.

\emph{Rule 1}:
    The generated shape correctly captured the constraint; only minor tweaks were needed. 

\emph{Rule 2}:
    The constraint logic was mostly correct. The InternalLink between elements had to be changed to include the inverse direction (since it is a directional connection in AutomationML and both directions are required).

\emph{Rule 3}:
    The target and constraint definition were mostly correct from the outset. The only fixes required were the additional definition of prefixes and the bidirectionality of InternalLinks.

The corrected \gls{SHACL} shapes were then applied to validate the \gls{AML} example model. The \gls{SHACL} validation process produces a validation report, formatted as a set of \gls{RDF} statements that specify which constraints were violated.

The \gls{LLM} was then prompted using a one-shot approach with both the \gls{SHACL} shapes and the resulting validation report. 
The prompt asked the model to generate a detailed explanation of the reported violations and to suggest concrete fixes for the underlying \gls{AML} model.
In all three cases, the \gls{LLM} correctly identified the cause of each violation and provided appropriate suggestions for resolving them. This demonstrates the approaches' capability not only to generate constraints but also to assist in understanding and repairing non-compliant data based on \gls{SHACL} validation output.
The \gls{SHACL}-shapes that were generated by the \gls{LLM}, the corrected versions, as well as the validation report and explanations, are available on GitHub\footnote{https://github.com/hsu-aut/aml-shacl}.

\subsection{Discussion}

The \gls{LLM}-driven pipeline presented in this paper produces \gls{SHACL} shapes that, while not immediately valid, required only minor adjustments from a domain expert. Much of this post-editing is a result of underspecified natural-language constraints rather than limitations of this approach. 
For instance, while \gls{AML} considers InternalLinks to be directional (e.g. an InternalLink from \textit{A} to \textit{B} is not equivalent to an InternalLink from \textit{B} to \textit{A}), the constraint implicitly assumes bidirectionality.

The experiment demonstrates that an \gls{LLM} can translate natural-language policy statements into near-complete \gls{SHACL} shapes with limited domain context. The remaining gap to correctness is small, enabling a human-in-the-loop workflow focused on validation rather than full specification.

However, the study also highlights a key limitation: natural-language rules often lack the precision needed for direct formalization. Future improvements will benefit most from clearer, more explicitly defined constraints.

Finally, once the \gls{SHACL} shapes were validated, the final step of our approach also proved useful in interpreting \gls{SHACL} validation reports and suggesting concrete fixes to the \gls{AML} data. 
In all cases, the model offered reasonable and accurate corrections, further reducing the maintenance effort.

In summary, the findings support a workflow where \glspl{LLM} generate \gls{SHACL} drafts and assist with debugging, while domain experts perform targeted reviews—achieving notable efficiency gains without compromising semantic accuracy.

\section{Conclusion}
\label{sec:conclusion}
This work-in-progress introduces a novel pipeline that leverages \glspl{LLM} to formalize textual constraints against \gls{AML}, e.g., from \glspl{AR}, into SHACL shapes, enabling automated validation of \gls{AML} models via semantic technologies. By transforming \gls{AML} models into \gls{OWL} ontologies and using \glspl{LLM} to both generate and interpret \gls{SHACL} constraints, the approach provides a human-in-the-loop solution that bridges the gap between informal documentation and formal model validation. An example use case from the \gls{AR} \emph{APC} illustrates that even complex domain-specific modeling rules can be verified and explained semi-automatically.

While initial results demonstrate the feasibility and effectiveness of the approach for selected recommendations, broader evaluation is needed. Future work should extend testing across a wider range of \glspl{AR} and more complex \gls{AML} models to ensure generalizability and robustness.

Importantly, the core idea of this approach is not limited to \gls{AML}. The methodology, i.e., \gls{LLM}-based \gls{SHACL} generation and validation on top of \gls{OWL}-based mappings, is adaptable to other industrial technologies such as the Asset Administration Shell or OPC UA. 
With suitable mappings from these technologies to \gls{OWL}, the same validation mechanism can be reused.
In particular, this opens up the potential for automated compliance checking of OPC UA Companion Specifications, providing a path toward semantic model verification across diverse industrial standards.

\bibliographystyle{IEEEtran}
\bibliography{references} 

\begin{thebibliography}{10}
\providecommand{\url}[1]{#1}
\csname url@samestyle\endcsname
\providecommand{\newblock}{\relax}
\providecommand{\bibinfo}[2]{#2}
\providecommand{\BIBentrySTDinterwordspacing}{\spaceskip=0pt\relax}
\providecommand{\BIBentryALTinterwordstretchfactor}{4}
\providecommand{\BIBentryALTinterwordspacing}{\spaceskip=\fontdimen2\font plus
\BIBentryALTinterwordstretchfactor\fontdimen3\font minus
  \fontdimen4\font\relax}
\providecommand{\BIBforeignlanguage}[2]{{%
\expandafter\ifx\csname l@#1\endcsname\relax
\typeout{** WARNING: IEEEtran.bst: No hyphenation pattern has been}%
\typeout{** loaded for the language `#1'. Using the pattern for}%
\typeout{** the default language instead.}%
\else
\language=\csname l@#1\endcsname
\fi
#2}}
\providecommand{\BIBdecl}{\relax}
\BIBdecl

\bibitem{IEC62714-1:2018}
{International Electrotechnical Commission}, ``{IEC 62714-1:2018: Engineering
  data exchange format for use in industrial automation systems engineering -
  Automation Markup Language - Part 1: Architecture and general
  requirements},'' IEC, Geneva, Switzerland, Standard, Apr. 2018.

\bibitem{Westermann2025}
\BIBentryALTinterwordspacing
T.~Westermann, M.~Ramonat, J.~Hujer, F.~Gehlhoff, and A.~Fay, ``{Automatic
  Mapping of AutomationML Files to Ontologies for Graph Queries and
  Validation},'' 2025. [Online]. Available:
  \url{https://arxiv.org/abs/2504.21694}
\BIBentrySTDinterwordspacing

\bibitem{Runde2009}
S.~Runde, K.~Güttel, and A.~Fay, ``Transformation of the caex plant design
  data in owl: An application of semantic web technologies in automation
  technology,'' \emph{VDI Berichte}, no. 2067, p. 175 – 178, 2009.

\bibitem{GrangelGonzalez.2016}
I.~Grangel-Gonz{\'a}lez, D.~Collarana, L.~Halilaj, S.~Lohmann, C.~Lange, M.-E.
  Vidal, and S.~Auer, ``{Alligator: A Deductive Approach for the Integration of
  Industry 4.0 Standards},'' in \emph{International Conference Knowledge
  Engineering and Knowledge Management}.\hskip 1em plus 0.5em minus 0.4em\relax
  {Springer}, 2016, pp. 272--287.

\bibitem{GrangelGonzalez.2018}
I.~Grangel-Gonz{\'a}lez, L.~Halilaj, M.-E. Vidal, O.~Rana, S.~Lohmann, S.~Auer,
  and A.~W. M{\"u}ller, ``Knowledge graphs for semantically integrating
  cyber-physical systems,'' in \emph{Database and Expert Systems Applications:
  29th International Conference, DEXA 2018}.\hskip 1em plus 0.5em minus
  0.4em\relax {Springer, Cham}, 2018, pp. 184--199.

\bibitem{OlgaKovalenko.2018}
\BIBentryALTinterwordspacing
{O. Kovalenko}, {I. Grangel-Gonz{\'a}lez}, {M. Sabou}, {A. L{\"u}der}, {S.
  Biffl}, and {S. Auer}, ``{AutomationML Ontology}: {Modeling Cyber-Physical
  Systems for Industry 4.0},'' \emph{Semantic Web}, vol. 2018, 2018. [Online].
  Available:
  \url{https://www.semantic-web-journal.net/system/files/swj1855.pdf}
\BIBentrySTDinterwordspacing

\bibitem{L.Abele.2015}
{L. Abele}, {C. Legat}, {S. Grimm}, and {A. W. M{\"u}ller}, ``Ontology-based
  validation of plant models,'' in \emph{2015 IEEE 13th International
  Conference on Industrial Informatics (INDIN)}, 2015, pp. 236--241.

\bibitem{LMM_AnLLMAidedEnterpriseKnowledge_2024}
E.~Laurenzi, A.~Mathys, and A.~Martin, ``{An LLM-Aided Enterprise Knowledge
  Graph (EKG) Engineering Process},'' \emph{{Proceedings of the AAAI Symposium
  Series}}, vol.~3, no.~1, pp. 148--156, 2024.

\bibitem{RaO_R2RMLChatGPTFramework_2024}
A.~Randles and D.~O'Sullivan, ``{R2 [RML]-ChatGPT Framework},'' in \emph{{5th
  International Workshop on Knowledge Graph Construction (KGCW 2024)}}, 2024.

\bibitem{ARAPC}
{AutomationML Association}, ``{N}ew {V}ersion of the {A}{R} {A}{P}{C},''
  \url{https://www.automationml.org/news/new-version-of-the-ar-apc-is-now-available/},
  2023.

\end{thebibliography}

\end{document}